\documentclass[letterpaper, 10 pt, conference]{ieeeconf}  

\usepackage{xcolor}
\usepackage{amsthm, amsmath}
\usepackage{amsfonts,amssymb,mathtools}
\usepackage{subcaption}
\usepackage{booktabs}
\usepackage{floatrow}
\usepackage{cleveref}

\newcommand{\OO}{\mathcal{O}}

\newcommand{\XX}{\mathcal{X}}

\newcommand{\real}{\mathbb{R}}

\newcommand{\st}{\textit{s.t. }}

\newcommand{\pd}[2][]{\frac{\partial#1}{\partial#2}}
\crefname{asp}{assumption}{assumptions}
\Crefname{asp}{Assumption}{Assumptions}
\crefname{lem}{lemma}{lemmas}
\Crefname{lem}{Lemma}{Lemmas}

\theoremstyle{definition}
\newtheorem{thm}{Theorem}
\newtheorem{lem}[thm]{Lemma}
\newtheorem{asp}[thm]{Assumption}

\newtheorem{defn}[thm]{Definition}

\theoremstyle{definition}
\newtheorem{case}{Case}

\theoremstyle{remark}

\theoremstyle{remark}

\usepackage[utf8]{inputenc}
\usepackage{comment}
\usepackage[normalem]{ulem}

\IEEEoverridecommandlockouts                              

\title{\LARGE \bf Zero-shot Transferable and Persistently Feasible Safe Control for\\ High Dimensional Systems by Consistent Abstraction
}
\author{Tianhao Wei$^{1}$, Shucheng Kang$^{2}$, Ruixuan Liu$^{1}$, and Changliu Liu$^{1}$
\thanks{*This material is based upon work supported by the National Science Foundation under Grant No. 2144489.}
\thanks{$^{1}$These authors are with Robotics Institute, Carnegie Mellon University, {\tt\small twei2, ruixuanl, cliu6@andrew.cmu.edu}}%
\thanks{$^{2}$Shucheng Kang is with the Department of Electrical Engineering, Tsinghua University. Work done during an internship at Carnegie Mellon.
        {\tt\small ksc19@mails.tsinghua.edu.cn}}%
}

\begin{document}

\maketitle
\thispagestyle{empty}
\pagestyle{empty}

\begin{abstract}                
Safety is critical in robotic tasks. Energy function based methods have been introduced to address the problem. To ensure safety in the presence of control limits, we need to design an energy function that results in persistently feasible safe control at all system states.
However, designing such an energy function for high-dimensional nonlinear systems remains challenging.
Considering the fact that there are redundant dynamics in high dimensional systems with respect to the safety specifications, this paper proposes a novel approach called abstract safe control.
We propose a system abstraction method that enables the design of energy functions on a low-dimensional model.
Then we can synthesize the energy function with respect to the low-dimensional model to ensure persistent feasibility.
The resulting safe controller can be directly transferred to other systems with the same abstraction, e.g., when a robot arm holds different tools. 
The proposed approach is demonstrated on a 7-DoF robot arm (14 states) both in simulation and real-world. Our method always finds feasible control and achieves zero safety violations in 500 trials on 5 different systems.
\end{abstract}



\section{Introduction}

Energy function-based methods have been extensively studied to ensure control-level safety for robotic systems in various applications, such as industrial robots in manufacturing or autonomous vehicles in transportation \cite{wei2019safe}. Typical approaches include the control barrier function method and the safe set algorithm (with safety index). These techniques aim to map dangerous states to high energy and safe states to low energy. Safety is guaranteed if a realizable control always exists that dissipates the energy whenever the state is in danger, which is known as \textit{persistent feasibility}.
Persistent feasibility can be guaranteed by offline energy function synthesis (known as the barrier function or the safety index)~\cite{wei2022safe}. Formal guarantees can be provided for general nonlinear systems with up to seven dimensions in states~\cite{bansal2017hamilton}.

However, existing energy function-based methods face challenges in ensuring persistent feasibility for general high-dimensional applications. For example, rule-based methods \cite{liu2014control} only apply for specific types of system dynamics; evolutionary optimization-based synthesis does not scale well for high dimensions \cite{wei2022safe} due to the curse of dimensionality; adversarial optimization lacks formal guarantees \cite{liu2022safe}; and Sum-of-Square-based methods are restricted to certain polynomial specifications \cite{clark2021verification, zhao2022safety}. Thus, methods to formally ensure safety for general high-dimensional applications are needed.

\begin{figure}[tb]
    \centering
    \includegraphics[width=\linewidth]{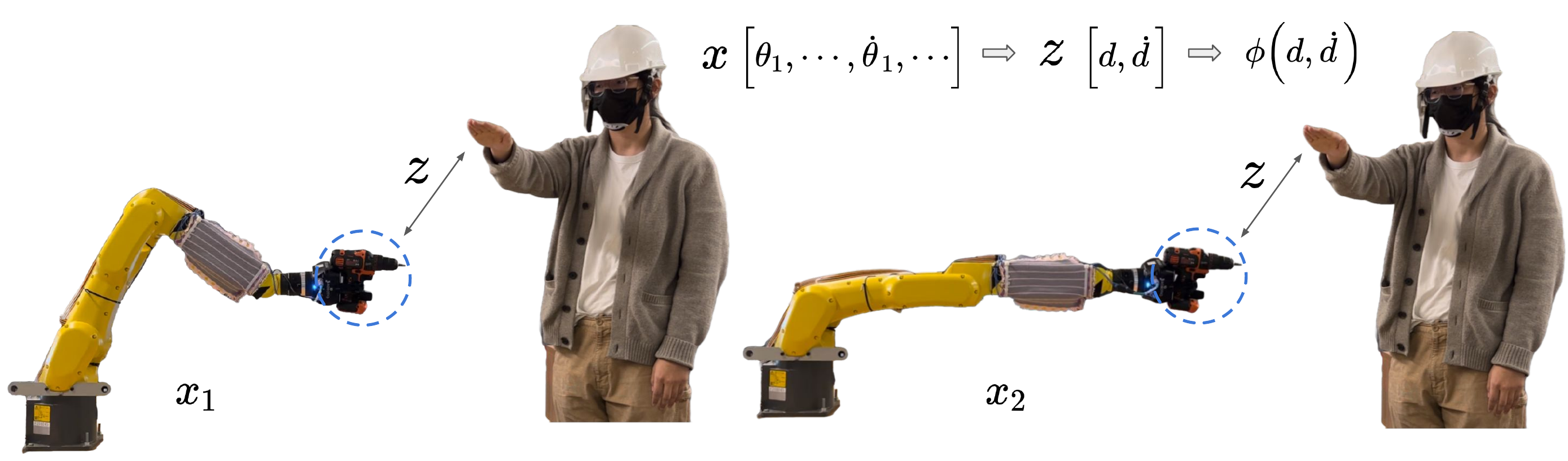}
    \caption{An illustration of abstract safe control to prevent collision between a drill and a human. The conventional safe control approach is to model the full kinematic chain of the robot system with state $x$. Our proposed approach considers abstract states $z=[d,\dot d]$ (relative distance and velocity) and a scalar M that considers constraints imposed by the full kinematic chain (whose values are different for $x_1$ and $x_2$).
    \vspace{-5mm}
    }
    \label{fig:idea}
\end{figure}

Our observation, as illustrated in \cref{fig:idea}, is that high-dimensional system models are typically redundant with respect to the safety specification. For instance, if we only need to ensure collision avoidance between a drill and a human, there is no need to verify the feasibility of safe control for all robot states. Similarly, when addressing collision avoidance for a legged robot, the focus is primarily on its center of mass rather than each leg's states~\cite{zhao2021model}. Hence, a simplified model of the robot can be used to replace the full dynamic model for safe control to reduce dimensionality.

A common approach to simplify dynamics is by employing a two-layer hierarchical architecture through system abstraction~\cite{pappas2000hierarchically}. The high-level system uses abstracted dynamics, while the low-level system uses the original concrete dynamics. 
However, existing system abstraction methods~\cite{smith2019continuous, yin2020optimization} can not ensure persistent feasibility. Because they usually assume the abstract model has uniform control constraints, which may lead to an unrealizable abstract control that breaks the feasibility. For instance, in \cref{fig:idea}, although both situations correspond to the same abstract state $z$, the feasible abstract control (Cartesian acceleration of the drill) in these two situations is different. In particular, at $x_2$, the horizontal acceleration of the drill is unrealizable due to singularity. 
To enable safe control through system abstraction, the notion of \textit{abstraction consistency} is essential to ensure that the high-level objective is always realizable by the concrete controller. Consistent abstraction of controllability~\cite{pappas2000hierarchically} and local accessibility \cite{pappas2002consistent} have been studied. However, it remains unclear on how to design a consistent abstraction for persistent feasibility, since the constraint is state-dependent, and control limits must be considered. 

To address these challenges, we propose a consistent abstraction for persistent feasibility, which allows us to design and verify the energy function (in the following discussion, we call it as the safety index) on a low-dimensional abstracted system. We specifically augment the abstract state with a scalar to account for different control constraints imposed by the concrete system for same abstract states.
For example, we extend $z$ in \cref{fig:idea} to $\hat z = [d, \dot d, M]$, where $M = \max |\ddot d|$. By mapping $x_1$ and $x_2$ to different $\hat z$, we can ensure that $\ddot d$ is always realizable by choosing it from the range $[-M,M]$. We will present a general method for designing this extended abstraction. Then we prove that a persistent feasible safety index synthesized on the extended abstraction guarantees persistent feasibility in the concrete system because all abstract controls are realizable. Lastly, we discuss how to design a persistent feasible safety index on the extended abstraction.

The abstraction not only reduces dimensionality but also enhances the transferability of the synthesized safety index. The safety index synthesized on such an extended abstraction can be directly applied to other concrete systems that have the same safety specifications (which implies same abstraction), as long as certain criteria is met. 
This transferability is especially useful for systems with time varying structures, such as a robot arm with changing end effectors. Once the safety index is feasible for a robot arm on the extended abstraction, it can be directly applied to the robot arm with different end effectors, to be shown in \cref{sec: experiment}.

\section{Formulation}

\subsection{Safety index synthesis}

Consider a control affine system~\footnote{Although our method assumes control affine dynamics, it is applicable to non-control affine systems, since we can always have a control affine form through dynamics extension~\cite{liu2016algorithmic}.}
\begin{align}
    (\Sigma_1) \quad & \dot x = f(x) + g(x) u\label{eq:x_dyn}
\end{align}
where $x \in X \subseteq \real^{n_x}$, $u \in U \subseteq \real^{n_u}$. We assume $U$ is a polytope, which is a common case in practice.
\begin{asp} [Polytope U]
    The control limits $U$ of the concrete system is a polytope: $U = \{u\mid A u \leq b\}$.
\end{asp}

A user-defined safety specification $\phi_0(x):\real^n \to \real$ is a continuous function and implicitly defines a connected and closed set $\XX_s := \{x\mid\phi_0(x)\leq0\}$ called the safe set. We are interested in keeping the state in a subset of the user-defined safe set: $S \subseteq \XX_s$. The problem can be expressed as a forward invariance problem: 
\begin{align}
   x(t_0) \in S \implies \forall t > t_0, x(t) \in S. 
\end{align}
In energy function-based methods, $S$ is defined by a designable safety index $\phi$: $S := \{x\mid\phi(x)\leq0\}$. The forward invariance can be guaranteed if the safe control constraint: $\dot \phi(x,u) < \gamma(x)$ is persistently feasible, where $\gamma(x)$ depends on the method. This paper considers the safe control constraint used in Safe Set Algorithm (SSA)~\cite{liu2014control}, corresponding to the following persistent feasibility condition:



\begin{defn}[Persistent feasibility]\label{def:x_feasibility}
    A safety index $\phi$ is persistently feasible if $\forall x \in X$ such that $\phi(x)=0$, there always exists $u\in U$ such that $\dot \phi(x,u) < 0$.
\end{defn}

However, designing a persistently feasible safety index to ensure forward invariance is not easy, especially for high-dimensional applications. Suppose $\phi$ is parameterized by $\theta$, the problem can be formulated as 
\begin{align}
    \min_{\theta} \left|B_{\theta}^*\right| := \min_{\theta} \left|\{x \mid \phi_{\theta}(x) = 0, \inf_{u}\dot\phi_{\theta}(x,u) \geq 0 \}\right|,\label{eq:B*}
\end{align}
where $B^*_\theta$ denotes the set of states on the boundary of $\phi_\theta$ that have no feasible safe control, $|B^*_\theta|$ is the volume of $B^*_\theta$. The goal is to optimize $\theta$ such that $\left|B^*_\theta\right|=0$. The task is difficult because computing $\left|B^*_\theta\right|$ is generally intractable for high dimensional systems.




\subsection{System abstraction}

To design a persistent feasible safety index for high-dimensional applications. We observe that, often, not all states are needed to check the satisfaction or feasibility of a safety specification as in \cref{eq:B*}. For instance, when considering collision avoidance of a tool held by a robot arm (\cref{fig:idea}), only the relative distance and velocity from the tool to the obstacle (2 dimensions) are required, instead of all 14 dimensions of the robot arm.

System $\Sigma_1$ is called the \textit{concrete system}. Suppose we want to design the safety index on a space $Z$ which is defined by a smooth, surjective map $z = \Phi(x)$, where $\Phi: \real^{n_x} \to \real^{n_z}$, $n_z \leq n_x$. Then we can define a system
\begin{align}
    (\Sigma_2) \quad & \dot z = f_z(z) + g_z(z) v\label{eq:z_dyn},
\end{align}
where $z \in Z \subseteq \real^{n_z}$, $v \in V \subseteq \real^{n_v}$. $v$ is an affine transformation of $u$, which will be presented in \cref{lem:affine_control}.
\begin{asp}
    We assume $g_z$ is of full column rank. Otherwise, the dimension of abstract safe control can be reduced. Therefore, $g_z(z)^{-1}$ always exists.
\end{asp}
\begin{defn}[$\Phi$-related and Abstraction] A system $\Sigma_2$ is $\Phi$-related to a system $\Sigma_1$ if for every trajectory $x(t)$, $z(t) = \Phi(x(t))$ is a trajectory of $\Sigma_2$. $\Sigma_2$ is an \textit{abstraction} of $\Sigma_1$ if it is $\Phi$-related to $\Sigma_1$~\cite{pappas2002consistent}.
\end{defn}
\cite{pappas2000hierarchically} proves that given a control system $\Sigma$ and any smooth map $\Phi$, there always exists a control system which is $\Phi$-related to $\Sigma$. \cite{pappas2002consistent} provides a method to construct the smallest $\Sigma_2$ on $Z$ that is $\Phi$-related to $\Sigma_1$. %
If $f_z(z)$ and $g_z(z)$ are unknown, they can be constructed with this method.


However, $\Phi$-relatedness is insufficient for designing the safety index on the abstraction. Because an abstract control at $z$ may not be implementable at all states $x = \Phi^{-1}(z)$ by the concrete system. For example, as in \cref{fig:idea}, a horizontal acceleration is implementable at $x_1$ but not at $x_2$. To enable designing safety index on the abstraction, we define consistent abstraction of constraint feasibility as follows:
\begin{defn}[Feasibility consistent abstraction]\label{def:con-abs}
    Let $\Sigma_1$ and $\Sigma_2$ be two control systems and $\Phi:X \to Z$ be a smooth map. Given a safety index $\phi$ defined on $X$. $\Sigma_2$ is a feasibility consistent abstraction of $\Sigma_1$ iff there exists a safety index $\phi_z: Z\mapsto \real$ such that the following conditions are satisfied: 1) $\phi_z(z) = \phi_z(\Phi(x)) =\phi(x)$; 2) $\forall z~ \st \phi_z(z)=0$, if $\exists v \in V, \st \dot{\phi_z}(z, v) < 0$, then  $\exists u \in U, \st \dot \phi(x, u) < 0$, $\forall x, \st \Phi(x) = z$.
\end{defn}

Consistent abstraction has been studied without considering control limits~\cite{pappas2002consistent, pappas2000hierarchically}. That is, when $U = \real^{n_u}$ and $V = \real^{n_v}$. But in reality, $U$ is usually a bounded set. How to design the abstraction under control limits has not been studied. Besides, the safety constraint $\dot \phi(x,u) < 0$ is state-dependent which introduces another difficulty. 


In the following, we first show how to construct feasibility-consistent abstractions under control limits, and characterize when persistent feasibility on the concrete system can be guaranteed by ensuring persistent feasibility on the abstracted system, that is, when condition 2) in \cref{def:con-abs} can be satisfied. Then we show how to ensure persistent feasibility on the abstracted system by designing a safety index.

\section{Consistent Abstraction Theory}




\subsection{Consistent abstraction under control limits}

In order to construct a consistent abstraction under control limits, we first show how to choose $\Phi$, the corresponding abstract space $Z$ and $\phi_z$. Then we pose a condition on the abstract control limits to make the abstraction consistent.

Suppose $\phi_0$ is composed of some inner functions. That is, $\exists \varphi_1(x),\cdots,\varphi_k(x)$, $\st \phi_0(x) = \phi_0(\varphi_1(x), \cdots, \varphi_k(x))$. For example, $\varphi_i(x)$ can be the distance to obstacles, the center of mass, etc. Inspired by \cite{liu2014control}, we can define $\Phi$ and correspondingly $Z$ by
\begin{align}
    \Phi = \varphi_1 \times \dot \varphi_1 \times \cdots \times \varphi_1^{(n_1)} \times \cdots \times \varphi_k \times \cdots \times \varphi_k^{(n_k)}
\end{align}
where the relative degree in the sense of Lie derivative from $\varphi_i^{(n_i)}$ to the concrete control $u$ is one. This choice of $\Phi$ ensures the appearance of $u$ in $\dot z$, which is a necessary condition of persistent feasibility. $z_i$ are all functions of $x$. 
Therefore, we can restrict the safety index $\phi(x)$ to be composite functions of $\Phi$ and let $\phi_z(z) := \phi(z)$. Then condition 1) in \cref{def:con-abs} is satisfied.
\begin{asp}[Composition]
    $\phi(x)$ is a composite function of $\Phi$, that is, $\phi: X\mapsto Z \mapsto \real$. We denote the mapping $Z\mapsto \real$ by $\phi(z)$ and have that $\phi(x) = \phi(\Phi(x)) = \phi(z)$. Furthermore, $\dot \phi(x,u) = \pd[\phi]{x}\dot x = \dot \phi(z,v) = \pd[\phi]{z}\dot z$
\end{asp}


For example, as shown in \cref{fig:idea}, we can define $\varphi(x) := d$, then $\phi_0(x) := 1-d = 1 - \varphi(x)$. We can let $z = [\varphi(x), \dot \varphi(x)] = [d, \dot d]$ and $\phi(z) = \phi(d, \dot d)$. 

Next, we present the condition of consistent abstraction. We start by proving that the abstract control $v$ is an affine transformation of the concrete control $u$.

\begin{lem}[Affine transformation of control]\label{lem:affine_control}
    Consider a control affine concrete system and a smooth map $\Phi$. The abstract control can be represented as an affine transformation of the concrete control, specifically, we can let $v = C(x) u + d(x)$, where $C(x) = g_z(\Phi(x))^{-1} \nabla \Phi(x) g(x)$, and $d(x) = g_z(\Phi(x))^{-1}[\nabla \Phi(x) f(x) - f_z(\Phi(x))]$. 
        \begin{proof}
    We have $\dot z = f_z(z) + g_z(z) v =  f_z(\Phi(x)) + g_z(\Phi(x)) v $ and $\dot z = \nabla \Phi(x) \dot x =  \nabla \Phi(x) f(x) + \nabla \Phi(x) g(x) u$.
    Rearrange the terms we will have $v =  g_z(\Phi(x))^{-1}[\nabla \Phi(x) f(x) - f_z(\Phi(x))] \nonumber
            + g_z(\Phi(x))^{-1} \nabla \Phi(x) g(x) u
        =   C(x) u + d(x)$
    \end{proof}
\end{lem}

\begin{lem}[Implementable abstract control]\label{defn:affine_limits}
    Given a polytope control limits $U$ and a state $x \in X$, based on \cref{lem:affine_control}, the implementable abstract control limits at $x$ is also a polytope set, defined as $\Phi_U(x):= \{C(x) u + d(x) \mid \forall u \in U\}$. We denote the linear bounds of $\Phi_U(x)$ by $\hat A$ and $\hat b$, that is $\Phi_U(x) = \{v \mid \hat A(x) v < \hat b(x)\}$.
\end{lem}


\begin{thm}[Feasibility consistency]\label{thm:feasibility-propagation}
    Let $\Sigma_1$ and $\Sigma_2$ be two control systems and $\Phi:X \to Z$ be a surjective smooth map. Suppose the abstract control limits $V$ is a state-dependent set, then condition 2) in \cref{def:con-abs} can be satisfied
    if
    \begin{align}
        \forall z \in Z,\ V(z) \subseteq \cap_{x\in \Phi^{-1}(z)} \Phi_U(x)\label{eq:feasibility-propagation}
    \end{align}
    \begin{proof}
    For arbitrary $x \in X$ such that $\phi(x)=0$, we can let $z = \Phi(x)$ and have $\phi(z) = 0$. If there exists a $v \in V(z)$ such that $\dot \phi(z,v) < 0$, from \cref{eq:feasibility-propagation}, we know this $v \in V(z) \subseteq \Phi_U(x)$. Therefore, based on \cref{defn:affine_limits}, $\exists u \in U$ such that $v = C(x) u + d(x)$. And this $u$ satisfies that $\dot \phi(x,u) = \pd[\phi]{x}\dot x = \pd[\phi]{z} \pd[z]{x} \dot x = \pd[\phi]{z}\dot z = \dot \phi(z,v) < 0$.
    \end{proof}
\end{thm}


\Cref{thm:feasibility-propagation} states that the feasibility can be propagated from $\Sigma_2$ to $\Sigma_1$ if all abstract control $v\in V(z)$ at a given $z$ is always implementable on the concrete system at arbitrary $x \in \Phi^{-1}(z)$. 
However, $V(z)$ is difficult to construct because $\Phi^{-1}(z)$ is difficult to compute. For example, $\Phi^{-1}(z)$ may correspond to all possible poses of a robot arm given an end-effector status $z$. Besides, $z = \Phi(x)$ may aggregate $x$ with vastly different abstract control limits $\Phi_U(x)$ into the same class, making $V(z)$ very tight and even empty. As shown in \cref{fig:idea}, if we design control using $V(z) = \Phi_U(x_1) \cap \Phi_U(x_2)$, we will lose horizontal acceleration for both cases. Therefore, we need a better method that is easy to construct and improves conservativeness.

\subsection{Non-conservative abstraction under control limits}


We propose an extended abstraction that only aggregates $x$ with similar abstract control limits $\Phi_U(x)$. 
We first define a one-dimensional under-approximation of $\Phi_U(x)$, then define the extended abstraction.


\begin{figure}
    \centering
    \includegraphics[width=\linewidth]{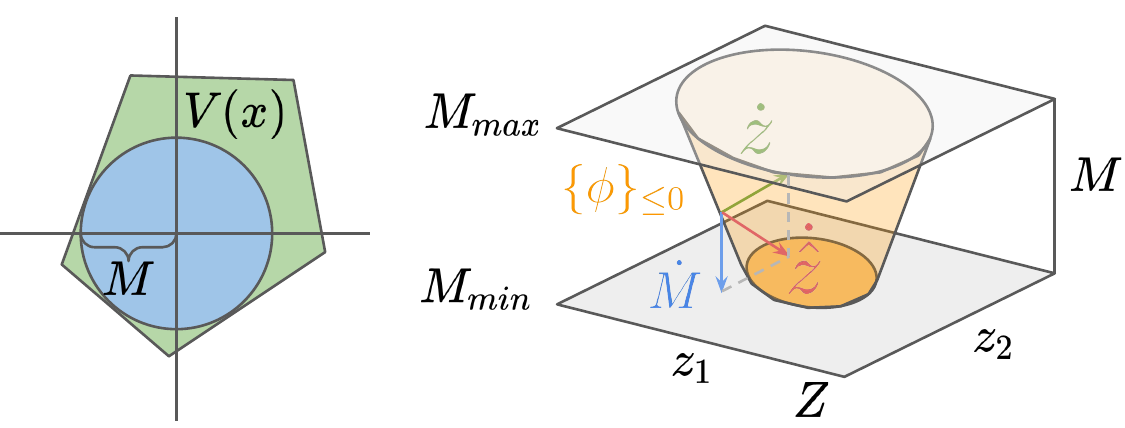}
    \caption{Left: Abstract control limit $V(x)$ and its under-approximation $M$. Right: Extended abstract safe set. The abstract space $Z$ is extended with the scalar $M$. 
    In this way, the abstract safe set is lifted from $Z$ to $\hat Z = Z \times \real$. Designing the safety index $\phi$ on $\hat Z$ enables different safe control under different control limits. Persistent feasibility requires that there always exists a control that leads to a system flow $\dot{\hat z}$ toward the interior of the invariance set. 
    \vspace{-5mm}
    }
    \label{fig:extended-abstraction}
\end{figure}

\begin{defn}[state-dependent radius of control constraints]
We define the radius of the largest zero-centered inner $Lp$-norm ball of $\Phi_U(x)$ as: 
\begin{align}
    M(x) = \max_{r} r \quad \st B_p(r) \subseteq \Phi_U(x),
\end{align}
where $B_p(r) = \{v \mid \|v\|_p \leq r\}$. 
\end{defn}

\begin{lem}[M computation]\label{lem:M}
$M(x)$ can be found by:
\begin{align}
    M = \max_{i}\ \hat b_i(x)/\|\hat a_i(x)\|_{p/(p-1)},
\end{align}
where $\hat a_i$ and $\hat b_i$ are rows of $\Phi_U(x)$'s linear boundary $\hat A$ and $\hat b$ defined in \cref{defn:affine_limits}. The proof is in \cref{apd:M_proof}.
\end{lem}

\begin{lem}[Bounded $\dot M$]\label{lem:bounded_dot_M}
    If $\hat a_i(x)$, $\hat b_i(x)$ are Lipschitz continuous and $\hat a_i(x)$ is non-zero, then $M(x)$ is Lipschitz continuous. Consequently, $\dot M := \lim_{dt \to 0} (M(x+\dot x dt)-M(x))/dt$ is bounded if $\dot x$ is bounded. 
    The proof is in \cref{apd:bounded_dot_M}.
\end{lem}

\begin{defn} [Extended abstraction]
    With $M(x)$, we define an extended abstraction for $\Sigma_1$ as $\hat \Phi(x) := \Phi(x) \times M(x)$, and define $\hat z := z \oplus M$. The extended abstract control limit is defined by $\hat V(\hat z) \subseteq B_p(M) \times \real$. The corresponding extended abstraction space is defined as $\hat Z := Z \times \real$, which has only one more dimension than $Z$.    
\end{defn}

In this way, we keep the extended abstraction in low dimension and gain flexibility in designing abstract control under different control limits.
One prerequisite of this under-approximation is that $\Phi_U(x)$ must contain $0$. Because $M(x) = 0$ when $0 \notin \Phi_U(x)$. In this case, $\hat V(\hat z)$ is an empty set which makes $\Sigma_2$ ill-defined. Therefore, we provide sufficient conditions of $0\in \Phi_U(x)$. 

\begin{asp}[$0 \in \Phi_U(x)$]
    We assume
    \begin{align}
        \forall x \in X, \exists u \in U,\ \st C(x)u + d(x) = 0.
    \end{align}{}
\end{asp}

To understand when the assumption holds, we present the following three case studies:
\begin{case}
    If $0\in U$ and systems are driftless~\cite{lavalle2006planning}, that is $f(x)=0$ and $f_z(z)=0$. Then $0 \in \Phi_U(x)$ always holds. 
\end{case}

\begin{case}
    When $v$ is a scalar, a sufficient condition of $\Phi_U(x)$ contains $0$ is $\min_{u\in \partial U} \|u\| > \max_{x \in X} \|d(x)\| / \|C(x)\|$.
\end{case}

\begin{case}
    When $v$ is a vector, a sufficient condition that $0 \in \Phi_U(x)$ is that $\min_{u\in \partial U} \|u\| > \max_{x \in X} \|C(x)^T w(x)\| \label{eq: vector_u}$,
    where $w(x)$ is a solution of $C(x) C(x)^T w(x) + d(x) = 0$.
\end{case}

The extended abstraction enables the design of abstract controllers under varying control limits. Without the extension, an abstract controller will give us the same control under different abstract control limits. But after the extension, we can define the safety index on $\hat Z$, which is abstract control limits aware. For example, as in \cref{fig:idea}, if the only considering $d,\dot d$, an abstract controller will suggest the same control for $x_1$ and $x_2$, but after the extension, it will suggest different actions for $x_1$ and $x_2$.

\section{Abstract Safe Control}
    
The consistent abstraction theory shows that a persistent feasible safety index defined on the extended abstraction guarantees persistent feasibility for the concrete system. Next, we show how to design a persistent feasible safety index on the extended abstract space $\hat Z$.

\subsection{Persistent feasibility on extended abstraction}

As shown in \cref{fig:extended-abstraction}, a safety index defined on $Z$ corresponds to a disk invariance set on the $Z$ plane. But a safety index defined on $\hat Z = Z \times \real$ corresponds to a bucket shape invariance set in $\hat Z$. $M$ can be viewed as an additional inner function. Persistent feasibility requires that we can always find a control on the boundary of the invariance set that leads to a system flow towards the inside of the invariant set. Formally, persistent feasibility requires the following inequality holds for all $x \in \{x\mid\phi(x)=0\}$:
\begin{align}
    & \min_{\|v\| < M} \dot \phi(x,u) = \dot \phi(z(x),v(u)) = \pd[\phi]{z} \dot z + \pd[\phi]{M} \dot M  \\
    & = \pd[\phi]{z(x)} [f_z(z(x)) + g_z(z(x)) v] + \pd[\phi]{M} \dot M(x) < 0 \label{eq:safe_X}
\end{align}

The numerical method to verify persistent feasibility for general nonlinear systems has an exponential growth time complexity $\OO(2^n)$~\cite{wei2022safe}, where $n$ is the dimension of the state. Therefore we wish to verify the feasibility on the abstracted system and propagate the feasibility back to concrete system. However, notice that $\dot M(x)$ does not depend on $z$ but on $x$. 
To guarantee the abstract control is consistent for all $x \in \Phi^{-1}(\hat z)$. We propose the following method. 

Suppose $M(x) \in [M_{min}, M_{max}] := R_v$, we instead verify the following stricter inequality: $\forall z\in\{z\mid\phi(z)=0\}, \forall M, \exists v$, such that
\begin{align}
    \min_{\|v\|<M} \pd[\phi]{z} \left[f_z(z) + g_z(z) v\right] + \left|\pd[\phi]{M} \dot M_{*}\right| < 0,\label{eq:stricter_feasibility}
\end{align}
where $\dot M_{*}$ is a chosen upper bound of $\dot M(x)$, which equals to or larger than the maximum value of $\dot M(x)$. It is easy to see that \cref{eq:stricter_feasibility} $\implies$ \cref{eq:safe_X}. Intuitively, \cref{eq:stricter_feasibility} says that we can always find a control $u$ that corresponds to a $\dot z$, whose combination with the largest possible $\dot M$ (which is $\dot{\hat z}$) still points towards the interior of the invariant set. And because \cref{eq:stricter_feasibility} is $x$-independent, we can verify it on the space $\hat Z$, which is a much smaller space than $X$, therefore can be verified by numerical methods. We can derive the following lemma from \cref{eq:stricter_feasibility}:\begin{lem}[Persistent feasibility]\label{lem:abstract_feasibility_M_max}
A safety index is persistently feasible if $\dot M_{*}$ satisfies the following condition for all $x\in X$ such that $\phi(x)=0$:
\begin{align}
    \inf_{z, M \in Z \times R_v} -\frac{\min_{\|v\|<M} \pd[\phi]{z} \left[f_z(z) + g_z(z) v\right]}{\left|\pd[\phi]{M}\right|} > \dot M_{*}. \nonumber
\end{align}
\end{lem}

In addition, \cref{eq:stricter_feasibility} can be relaxed when $M=M_{min}$. Because $\pd[\phi]{M}$ is expected to be always negative (the system is safer when the control limit is larger), and $\dot M \geq 0$ when $M=M_{min}$. Therefore $\pd[\phi]{M}|_{M_{min}} \dot M|_{M_{min}} \leq 0$. Therefore, we only need to verify that $\min_{\|v\|<M} \pd[\phi]{z} \left[f_z(z) + g_z(z) v\right]  < 0$ when $M = M_{min}$. We can estimate $M_{min}$ and $M_{max}$ by sampling the state space and computing $M$ with \cref{lem:M}.





\begin{lem}[Sampling guarantee]\label{lem:sampling}
    Suppose 1. we uniformly sample $N$ times from $X$ and denote the maximum and minimum of $M(x)$ by $\hat M_{max}$ and $\hat M_{min}$, and the maximum of $\dot M(x)$ by $\hat{\dot M}_{*}$; 2. a $\phi$ is designed to guarantee persistent feasibility with $\hat M_{max}$, $\hat M_{min}$ and $\hat{\dot M}_{*}$. Then we can guarantee that more than $100\cdot p\%$ states are feasible with the probability $[1-p^N]^3$, where $0<p<1$. The proof is in \cref{apd:sampling_proof}
\end{lem}

\subsection{Safety Index Synthesis and Execution}


With dynamics abstraction, we can synthesize $\phi$ on a low dimensional space. Therefore, we can use analytical or numerical methods~\cite{wei2022safe} to compute $\left|B^*_\theta\right|$ in \cref{eq:B*}.
During online execution, we can find a safe control in the original space when $\phi(x)=0$ by:
\begin{align}
    & \dot \phi(x, u) = \pd[\phi]{z} \pd[z]{x} \dot x + \pd[\phi]{M} \pd[M]{x} \dot x\\
    & = (\pd[\phi]{z} \pd[z]{x} + \pd[\phi]{M} \pd[M]{x}) [f(x) + g(x) u] < 0.
\end{align}
$\pd[M]{x}$ is sometimes difficult to derive analytically, but it can always be computed numerically by taking the limit: $\lim_{\delta \to 0} [M(x+\delta)-M(x)]/\delta.$ The numerical error can be compensated by using a larger ${\dot M}_*$.
With this safety constraint, we can choose a control $u$ that is close to a given safety-ignorant reference control $u_0$ with a QP objective function as in most energy function based method~\cite{wei2019safe}: $\min_{u} \|u-u_0\|\ \st \dot \phi(x,u) < 0.$

\section{Experiment}\label{sec: experiment}
    Experiments are designed to show that 1. the prerequisite of the method can be easily achieved; 2. the method ensures persistent feasibility for high dimensional systems; 3. the synthesized safety index can be transferred to systems with different dynamics.

The robot's task is to reach a goal while avoiding collision with obstacles or humans. We consider a collision avoidance constraint $\phi_0 = d_{min} - d$, where $d_{min}=0.05$ and $d$ is the relative distance.
Our methods are tested on a 7 degrees-of-freedom (DoF) Franka Panda robot arm in Panda3D simulation with a 1.5GHz AMD EPYC 7H12 64-Core Processor, and a 7 DoF FANUC LR Mate 200i real robot.

\subsection{Distribution of $M$ and $\dot M$}

\begin{figure}[tbh]
     \centering
     \begin{subfigure}[b]{0.48\linewidth}
         \centering
         \includegraphics[width=\textwidth]{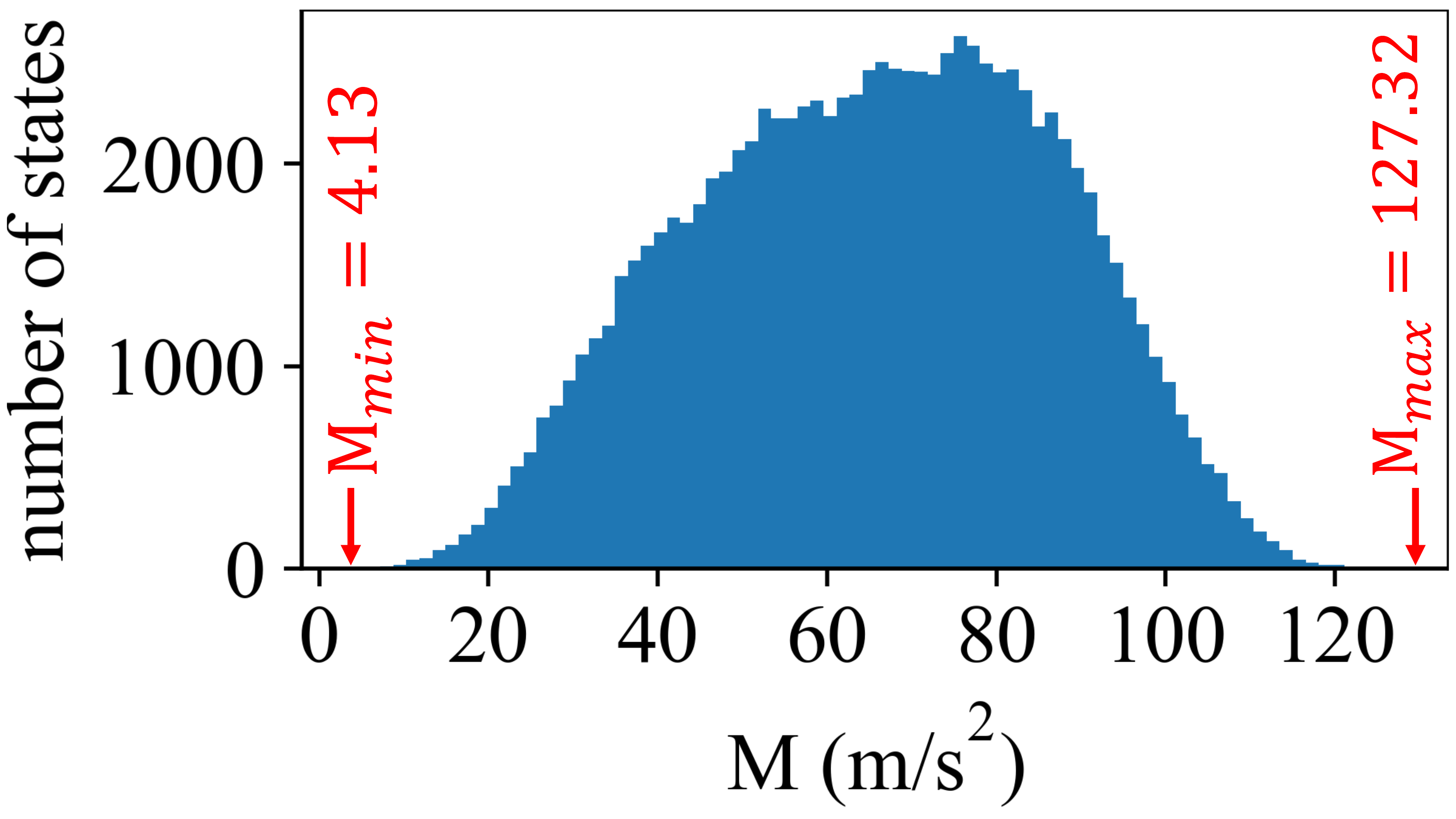}
         \caption{$M$ distribution. \vspace{-2mm}}
         \label{fig:M-dist}
     \end{subfigure}
     \hfill
     \begin{subfigure}[b]{0.48\linewidth}
         \centering
         \includegraphics[width=\textwidth]{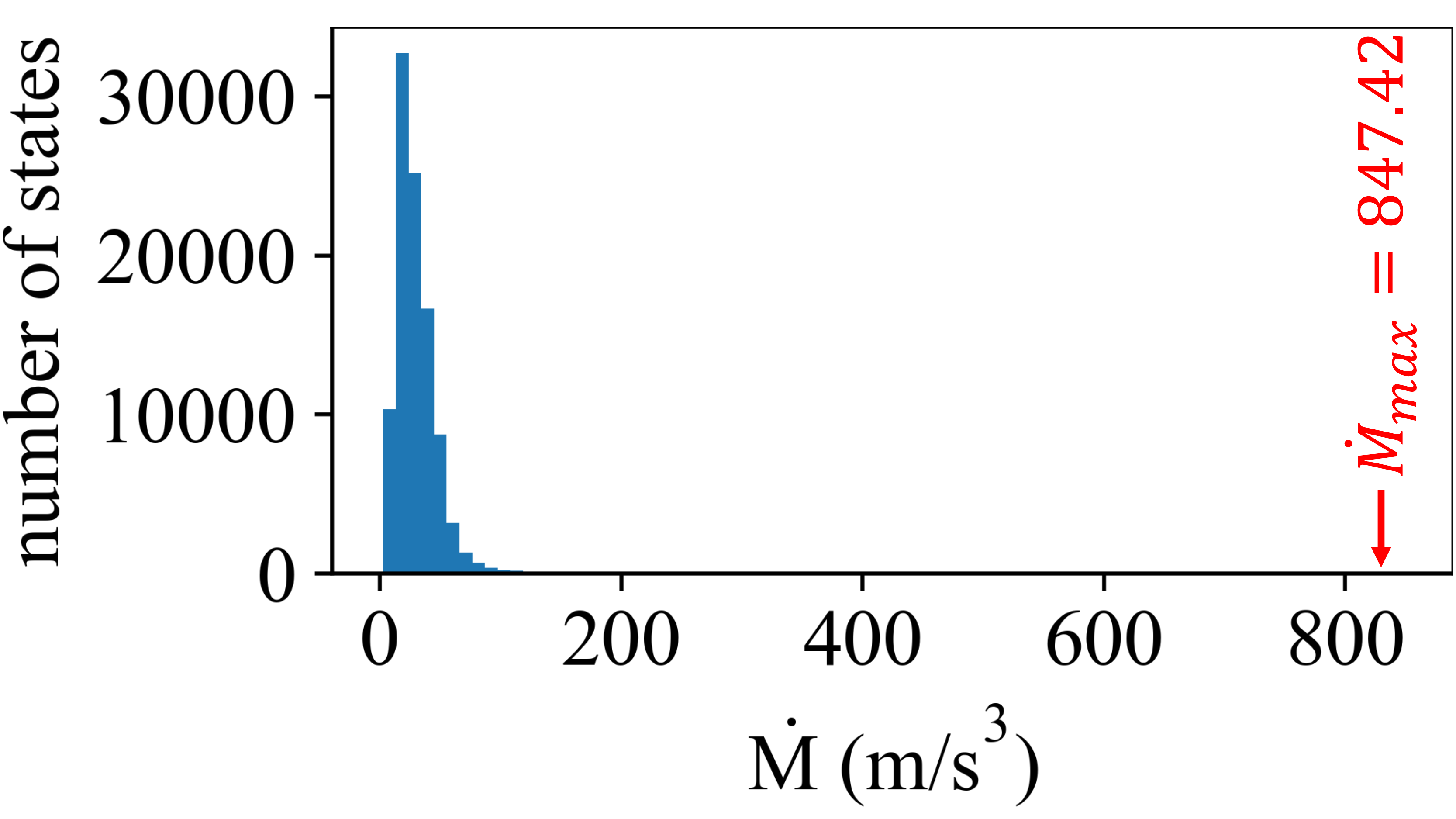}
         \caption{$\dot M$ distribution. \vspace{-2mm}}
         \label{fig:dot-M-dist}
     \end{subfigure}
    \caption{Distribution of $M$ and $\dot M$. \vspace{-4mm}}
    \label{fig:distribution}
\end{figure}

\begin{figure}[tbh]
     \centering
     \begin{subfigure}[b]{0.35\linewidth}
         \centering
         \includegraphics[height=1.5cm]{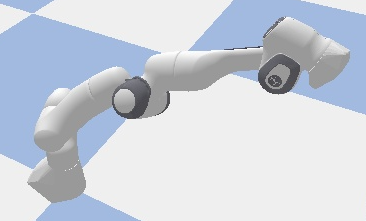}
         \caption{$\hat M_{min}$ pose\vspace{-2mm}}
         \label{fig:M-min}
     \end{subfigure}
     \hfill
     \begin{subfigure}[b]{0.3\linewidth}
         \centering
         \includegraphics[height=1.5cm]{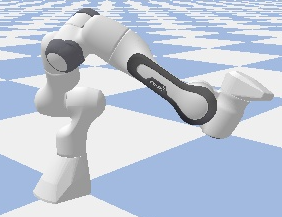}
         \caption{$\hat M_{max}$ pose\vspace{-2mm}}
         \label{fig:M-max}
     \end{subfigure}
     \hfill
     \begin{subfigure}[b]{0.25\linewidth}
         \centering
         \includegraphics[height=1.5cm]{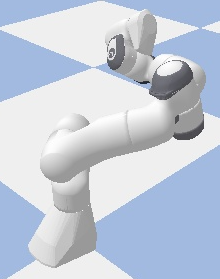}
         \caption{$\hat{\dot M}_*$ pose\vspace{-2mm}}
         \label{fig:dot-M-max}
     \end{subfigure}
    \caption{Poses for $\hat M_{max}$ $\hat M_{min}$ and $\hat{\dot M}_*$. \vspace{-4mm}}
    \label{fig:poses}
\end{figure}

We first show the distribution of $M$ and $\dot M$ in \cref{fig:distribution}. It reveals that $M$ is always above $0$, very few states have a small range, and Therefore, the extended abstraction greatly reduced conservativeness.
Most states have a small $\dot M$ which eases finding feasible safe control. It takes 14 hours to sample 100000 $\dot M$ and estimate $\dot M_{*}$.
\Cref{fig:poses} shows the corresponding poses for $\hat M_{max}$ $\hat M_{min}$ and $\hat{\dot M}_*$. We can see that these poses usually corresponds to singular states, such as fully extended. Therefore, users may accelerate the system property verification by starting from these poses.

\subsection{Persistent Feasibility}

The safety index is designed as the following form:
\begin{align}
    \phi = \max(\phi_0, \phi^*),\ \text{where }
    \phi^* = d_{min}^2-d^2-k\dot{d}/M.
\end{align}
The persistent feasibility of $\phi$ can be guaranteed by the persistent feasibility of $\phi^*$ as shown in~\cite{liu2014control}. Therefore, we can use system abstraction to directly determine the value of $k$ that guarantees feasibility of $\phi^*$ and $\phi$. Assuming $d\in [0.0, 0.8] m$ and $\dot{d}\in [-1.0, 1.0] m/s$: 
$\forall (d, \dot{d}, M)$, s.t. $\phi^*=0$, there exists $\ddot{d}\in \left[ -M,M \right]$ such that:  
\begin{align}
    \dot{\phi}^* &= -2d\dot{d}-k\frac{\ddot{d}}{M}+k\frac{\dot{d}}{M^2}\dot{M} \\
    &\le -2d\dot{d}-k\frac{\ddot{d}}{M}+\mid \dot{d}\mid \frac{k}{M^2}\dot{M}_{\max}\le 0
\end{align}
We let $\ddot{d}=M$, then $k\ge -2d\dot{d}+ k|\dot{d}| \dot{M}_{\max} /M^2$.
Since $\phi^*=0$, we have $k=M(d_{min}^{2}-d^2)/\dot{d}$ and 
\begin{align}
    k &\ge -2d\dot{d}+\frac{\mid \dot{d}\mid}{\dot{d}}\left( d_{min}^{2}-d^2 \right) \cdot \frac{\dot{M}_{\max}}{M} \\
    &\ge 2\mid d\dot{d}\mid _{\max}+\mid d_{min}^{2}-d^2\mid _{\max}\cdot \frac{\dot{M}_{\max}}{M_{\min}}
\end{align}
which implies $k \ge 133.31$. Based on \cref{lem:sampling}, we can guarantee that at least $99.99\%$ states are feasible with the probability $99.99\%$.
As shown in Table \ref{tab:results}, this value of $k$ ensures no collisions in 100 randomly generated goal-obstacle pairs. 
In contrast, it is too complex to derive the analytical condition for the concrete system, which has hundreds of terms with nonlinear transformations such as trigonometric functions and multiplication.
The exponential time complexity of numerical methods also makes it infeasible to check feasibility. For example, for a 7 degree-of-freedom robot arm, even with only 10 samples per dimension, it would require checking $10^{14}$ samples.

\begin{table}[tb]
    \caption{Collision count in 100 randomly generated scenarios and dynamics sensitivity to DoF and End Effectors. \vspace{-5mm}}
    \centering
    \small
    \begin{tabular}{cccc}
        \toprule
        Case         & Collision & Range of $M$         & $\dot M_{*}$ \\
        \midrule
        5 DoF + EE 1 & 0         & {[1.67, 114.37]} & 1052.93       \\
        6 DoF + EE 1 & 0         & {[3.26, 121.34]} & 891.80         \\
        7 DoF + EE 1 & 0         & {[4.13, 127.32]} & 847.42         \\
        7 DoF + EE 2 & 0         & {[4.53, 120.19]} & 710.97         \\
        7 DoF + EE 3 & 0         & {[3.82, 131.55]} & 856.18         \\
        \bottomrule
        \end{tabular}\label{tab:results}
\end{table}

\subsection{Transplant to other systems}

In this experiment, we change robot's degree-of-freedoms (DoF) and end-effectors (EE) to discuss when the safety index can be directly transferred to other systems. EE 1 is the baseline. In EE 2, a $0.1kg$-rod is attached to the end-effector. In EE 3, the offset of the seventh joint has a 0.05m increment in y-axis. In 6 DOF robot, the seventh joint is fixed. In the 5 DOF robot, both the sixth and seventh joints are fixed. As shown in \cref{tab:results}, $\dot M_{*}$ decreases when the DoF increases, and the range of $M$ does not change too much with the dynamics. We can conclude that if we consider a small enough of $M_{min}$ and a large enough $\dot M_{*}$ during the design of the safety index, the same safety index can ensure safety for a wide range of unseen systems. This method is particularly useful for robot arms with real-time tool switching.

\subsection{Experiment on real robots}

\begin{figure}
    \centering
    \includegraphics[width=.8\linewidth]{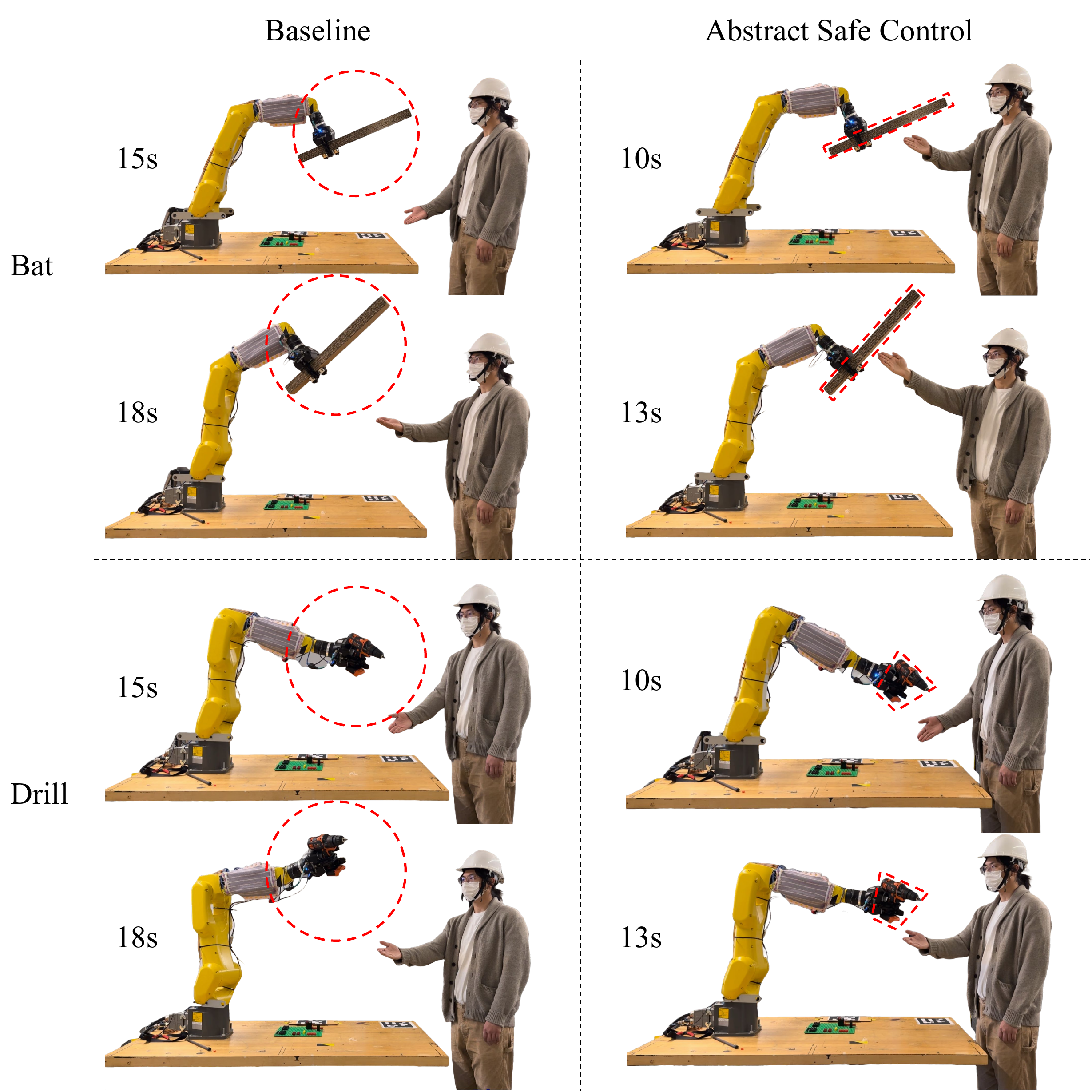}
    \caption{Experiments on FANUC with two different tools. The baseline method ensures safety by considering a large sphere that contains all possible tools. But our method can guarantee safety by considering the nearest point from the tool to the human, which is much less conservative than the baseline.}
    \label{fig:FANUC_exp}
\end{figure}

\begin{figure}[htb]
\centering
\begin{subfigure}{0.45\textwidth}
\includegraphics[width=\textwidth]{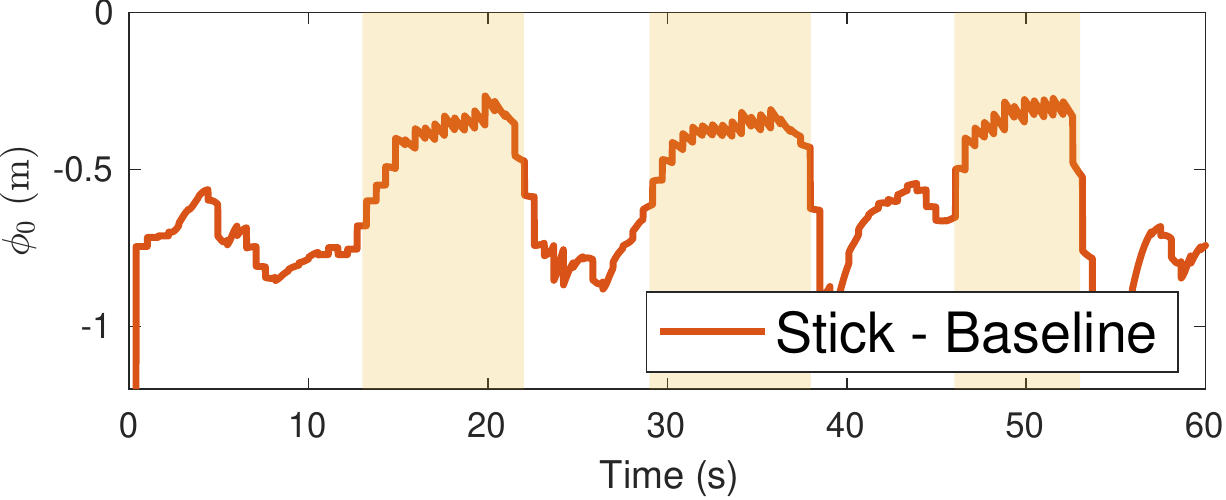}
\end{subfigure}
\begin{subfigure}{0.45\textwidth}
\includegraphics[width=\textwidth]{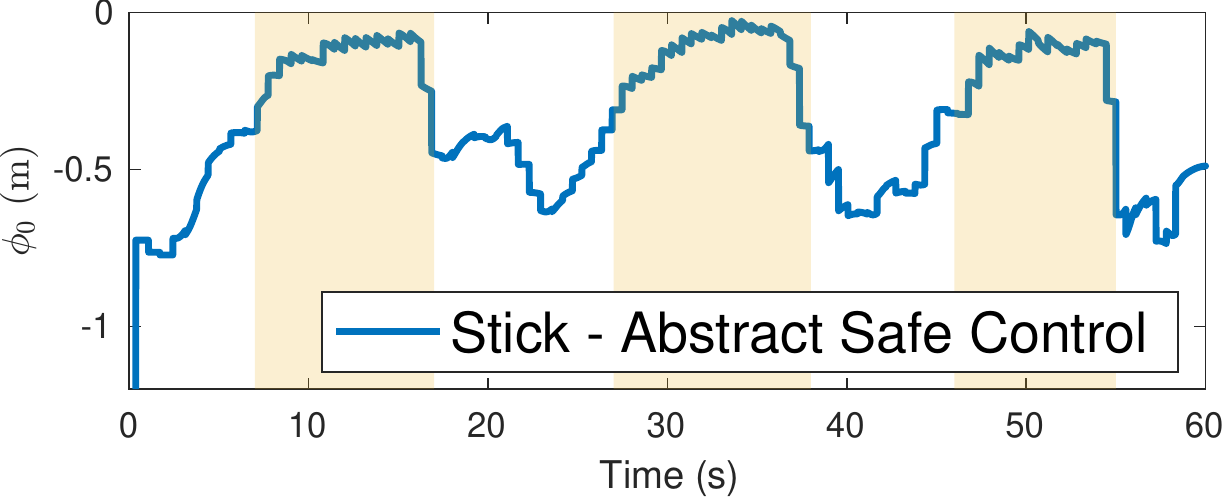}
\end{subfigure}
\par\medskip
\begin{subfigure}{0.45\textwidth}
\includegraphics[width=\textwidth]{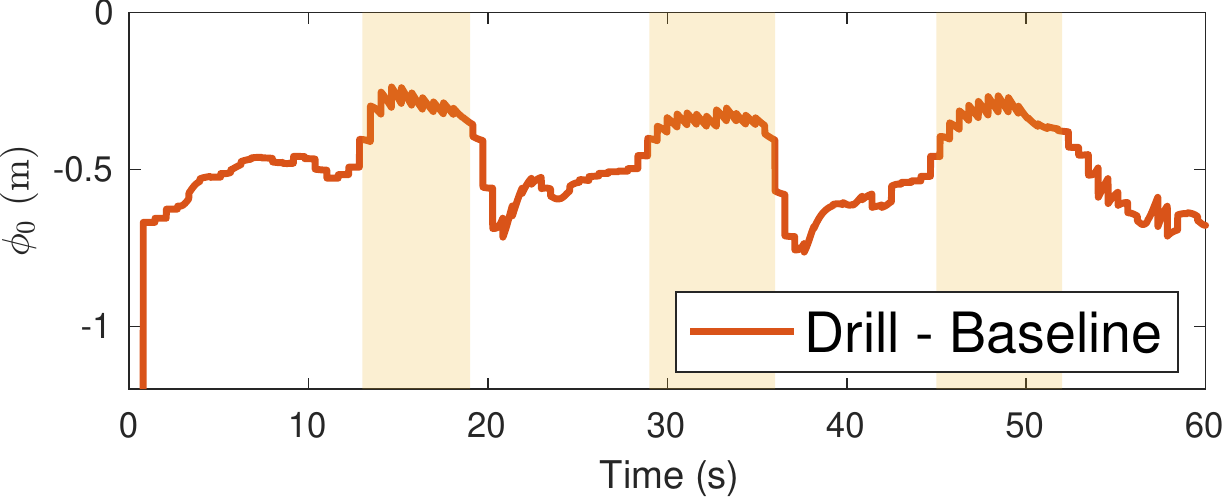}
\end{subfigure}
\begin{subfigure}{0.45\textwidth}
\includegraphics[width=\textwidth]{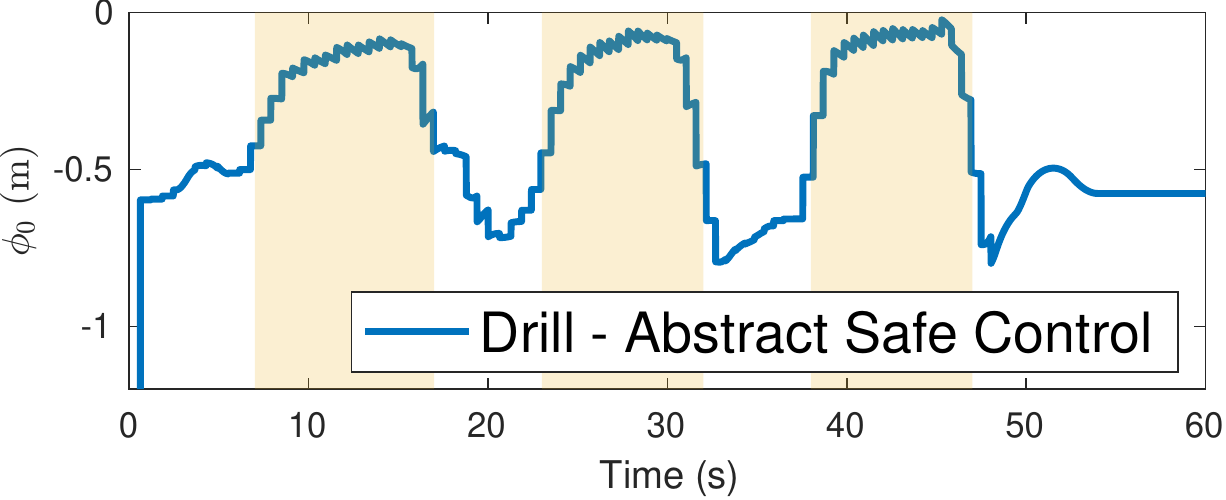}
\end{subfigure}
\caption{$\phi_0$ profiles ($-d$). Our abstract safe control method is much less conservative, allowing a closer distance to the human while maintaining safety. The interaction sessions are shaded in yellow. \vspace{-5mm}}
\label{fig:FANUC_profile}
\end{figure}

We also test the method on a real FANUC LR Mate 200i robot with two different tools: a drill and a bat. These two tools have different shapes and different kinematics. The relative distance/velocity is estimated by detecting the human skeleton with a Kinect Camera in real time. Our method significantly reduces conservativeness when the tool is unknown. As shown in \cref{fig:FANUC_exp}, to ensure safety during tool use, the baseline method constructs a large sphere space that covers all possible tools. However, this method leads to conservative interaction with humans. But with our method, we can consider the nearest point from the robot to the human because the nearest points can be viewed as different end effectors that our safety index can be directly transplanted. Therefore  we can provide safety guarantees for both tools while remaining non-conservative, the profiles of $\phi_0$ are shown in \cref{fig:FANUC_profile}. 

\bibliographystyle{IEEEtran}
\bibliography{reference}             
                                                   







\appendix

\subsection{\Cref{lem:M} computing $M$ proof}\label{apd:M_proof}
\begin{proof}
The radius of the maximum $L_p$-norm inner ball can be found by
\begin{align}
    \max_{r} r \quad \st \hat a_i(x) v \leq \hat b_i(x), \forall \|v\|_p \leq r.
\end{align}
Based on Hölder's inequality, for $p,q$ that satisfy  $\frac{1}{p}+\frac{1}{q}=1$, we have
\begin{align}
    |\hat a_i(x) v| \leq \|\hat a_i(x)\|_q \|v\|_p
\end{align}
Therefore
\begin{align}
    \max_{\|v\|_p\leq r} \hat a_i(x) v = \max_{\|v\|_p\leq r} |\hat a_i(x) v| \leq \max_{M} \|\hat a_i(x)\|_q M
\end{align}
\end{proof}

\subsection{\Cref{lem:bounded_dot_M} Bounded $\dot M$ proof}\label{apd:bounded_dot_M}
\begin{proof}
    Based on \cref{lem:M}, $M$ is the maximum of finite Lipschitz continuous functions . Therefore, $M$ is also Lipschitz continuous. That means $(M(x+\delta)-M(x))/\|\delta\|$ is bounded by the Lipschitz constant $L$. Then 
    \begin{align}
        \dot M &= \lim_{dt \to 0} (M(x+\dot x dt)-M(x))/dt\\
        &= \left[ \lim_{dt \to 0} (M(x+\dot x dt)-M(x))/(\|\dot x\| dt)\right]\|\dot x\|\\
        &\leq \left[ \lim_{\delta \to 0} (M(x+\delta)-M(x))/\|\delta\|\right]\|\dot x \|\\
        &\leq L \|\dot x\|
    \end{align}
     is also bounded when $\dot x$ is bounded. Note that $M$ does not have to be differentiable.
\end{proof}

\subsection{\Cref{lem:sampling} sampling guarantee proof}\label{apd:sampling_proof}
\begin{proof}
    The probability of $N$ samples all being smaller than the $p$-quantile value is $p^N$. Therefore,  the probability of $\hat M_{max}$ being larger than $p$ quantile threshold is $1-p^N$. The same holds for $\hat M_{min}$ and $\hat{\dot M}_{*}$. Therefore, we can say that more than $100\cdot p\%$ states has a larger $M$ than $\hat M_{min}$, a smaller $M$ than $\hat M_{max}$ and a smaller $\dot M$ than $\hat{\dot M}_{*}$ with the probability $[1-p^N]^3$. These states all have feasible safe control with a safety index designed with $\hat M_{min}$, $\hat M_{max}$ and $\hat{\dot M}_{*}$.
\end{proof}

\subsection{Parameter Estimation in safety Index}                             
We just need to guarantee that $\dot{\phi} \le 0$, $\forall \phi=0$. This means for all $(d, \dot{d}, M)$, s.t. $\phi =d_{min}^{2}-d^2-k\frac{\dot{d}}{M}=0$, there exists $\ddot{d}\in \left[ -M,M \right]$ to satisfy the following inequality:  
\begin{align}
    \dot{\phi} &= -2d\dot{d}-k\frac{\ddot{d}}{M}+k\frac{\dot{d}}{M^2}\dot{M} \\
    &\le -2d\dot{d}-k\frac{\ddot{d}}{M}+\mid \dot{d}\mid \frac{k}{M^2}\dot{M}_{\max}\le 0
\end{align}
We let $\ddot{d}=M$, then $k\ge -2d\dot{d}+\mid \dot{d}\mid \frac{k}{M^2}\dot{M}_{\max}$.
Since $\phi=0$, we have $k=M\frac{d_{min}^{2}-d^2}{\dot{d}}$ and 
\begin{align}
    k &\ge -2d\dot{d}+\frac{\mid \dot{d}\mid}{\dot{d}}\left( d_{min}^{2}-d^2 \right) \cdot \frac{\dot{M}_{\max}}{M} \\
    &\ge 2\mid d\dot{d}\mid _{\max}+\mid d_{min}^{2}-d^2\mid _{\max}\cdot \frac{\dot{M}_{\max}}{M_{\min}}
\end{align}
which implies $k \ge 133.31$.

\end{document}